\documentclass[letterpaper, 10 pt, journal, twoside]{IEEEtran}  

\usepackage{graphicx}
\usepackage{subcaption}
\usepackage[framemethod=TikZ]{mdframed}
\usepackage{lipsum} 
\usepackage{caption}
\usepackage{wrapfig}
\usepackage{listings}
\usepackage{xcolor}
\usepackage{multirow}
\usepackage{makecell}
\usepackage[misc]{ifsym}
\usepackage{tikz}
\usepackage{fancyhdr}

\newif\ifhighlight
\highlightfalse  

\newcommand{\highlight}[1]{\ifhighlight\textcolor{red}{#1}\else#1\fi}

\lstdefinestyle{pythonstyle}{
    language=Python,                     
    basicstyle=\ttfamily\footnotesize,  
    numberstyle=\tiny,                  
    stepnumber=1,                       
    numbersep=5pt,                      
    backgroundcolor=\color{white},      
    showspaces=false,                   
    showstringspaces=false,             
    showtabs=false,                     
    tabsize=2,                          
    captionpos=b,                       
    breaklines=true,                    
    breakatwhitespace=true,             
    commentstyle=\color{green},         
    keywordstyle=\color{blue},          
    stringstyle=\color{red}             
}

\usepackage{amsmath,amsfonts,bm}

\def\eqref#1{equation~\ref{#1}}

\def\1{\bm{1}}

\DeclareMathAlphabet{\mathsfit}{\encodingdefault}{\sfdefault}{m}{sl}
\SetMathAlphabet{\mathsfit}{bold}{\encodingdefault}{\sfdefault}{bx}{n}

\DeclareMathOperator*{\argmax}{arg\,max}

\usepackage{color}
\usepackage{xcolor}
\definecolor{turquoise}{cmyk}{0.65,0,0.1,0.3}
\definecolor{purple}{rgb}{0.65,0,0.65}
\definecolor{dark_green}{rgb}{0, 0.5, 0}
\definecolor{orange}{rgb}{0.8, 0.6, 0.2}
\definecolor{red}{rgb}{0.8, 0.2, 0.2}
\definecolor{darkred}{rgb}{0.6, 0.1, 0.05}
\definecolor{blueish}{rgb}{0.0, 0.3, .6}
\definecolor{light_gray}{rgb}{0.7, 0.7, .7}
\definecolor{pink}{rgb}{1, 0, 1}
\definecolor{greyblue}{rgb}{0.25, 0.25, 1}
\definecolor{tab_blue}{HTML}{1f77b4}
\definecolor{tab_orange}{HTML}{ff7f0e}
\definecolor{LightRed}{rgb}{0.99,0.89,0.89}

\definecolor{mesh_misty_rose}{HTML}{e6aaa3}
\definecolor{mesh_yellow}{HTML}{ffba00}

\definecolor{our_red}{rgb}{0.99,0.89,0.89}
\definecolor{our_blue}{HTML}{1f77b4}
\definecolor{our_orange}{HTML}{ff7f0e}

\usepackage{hyperref}
\usepackage{url}
\usepackage{booktabs}
\usepackage{algorithm,algorithmicx,algpseudocode}
\algnewcommand{\algorithmicforeach}{\textbf{for each}}
\algdef{SE}[FOR]{ForEach}{EndForEach}[1]
  {\algorithmicforeach\ #1\ \algorithmicdo}
  {\algorithmicend\ \algorithmicforeach}

\begin{document}

\title{Make a Donut: Hierarchical EMD-Space Planning\\ for Zero-Shot Deformable Manipulation with Tools}

\author{Yang You$^{1,*}$, Bokui Shen$^1$, Congyue Deng$^1$, Haoran Geng$^2$, Songlin Wei$^2$, He Wang$^2$, Leonidas Guibas$^{1,*}$\\
\thanks{Manuscript received: September 8, 2024; Revised December 10, 2024; Accepted January 17, 2025.}
\thanks{This paper was recommended for publication by Editor Júlia Borràs Sol upon evaluation of the Associate Editor and Reviewers' comments.}
\thanks{$^{1}$Yang You, Bokui Shen, Congyue Deng, Leonidas Guibas are with Department of Computer Science, Stanford University, the US.
        }
\thanks{$^{2}$Haoran Geng, Songlin Wei, He Wang are with CFCS, Peking University, China.}
\thanks{*Contact email: \tt\small \{yangyou,guibas\}@stanford.edu}
\thanks{The research is supported in part by the Toyota Research Institute University 2.0 Program and a Vannevar Bush Faculty Fellowship. Yang You is also supported in part by the Outstanding Doctoral Graduates Development Scholarship of Shanghai Jiao Tong University.}
\thanks{Digital Object Identifier (DOI): 10.1109/LRA.2025.3537899}
}

\markboth{IEEE Robotics and Automation Letters. Preprint Version. Accepted January, 2025}
{You \MakeLowercase{\textit{et al.}}: Make a Donut: Hierarchical EMD-Space Planning for Zero-Shot
Deformable Manipulation with Tools} 

\maketitle


\begin{abstract}
Deformable object manipulation stands as one of the most captivating yet formidable challenges in robotics. While previous techniques have predominantly relied on learning latent dynamics through demonstrations, typically represented as either particles or images, there exists a pertinent limitation: acquiring suitable demonstrations, especially for long-horizon tasks, can be elusive. Moreover, basing learning entirely on demonstrations can hamper the model's ability to generalize beyond the demonstrated tasks. In this work, we introduce a demonstration-free hierarchical planning approach capable of tackling intricate long-horizon deformable manipulation tasks without necessitating any training. We employ large language models (LLMs) to articulate a high-level, stage-by-stage plan corresponding to a specified task. For every individual stage, the LLM provides both the tool's name and the Python code to craft intermediate subgoal point clouds. With the tool and subgoal for a particular stage at our disposal, we present a granular closed-loop model predictive control strategy. This leverages Differentiable Physics with Point-to-Point correspondence (DiffPhysics-P2P) loss in the Earth Mover Distance (EMD) space, applied iteratively. Experimental findings affirm that our technique surpasses multiple benchmarks in dough manipulation, spanning both short and long horizons. Remarkably, our model demonstrates robust generalization capabilities to novel and previously unencountered complex tasks without any preliminary demonstrations. We further substantiate our approach with experimental trials on real-world robotic platforms. Our project page: \url{https://qq456cvb.github.io/projects/donut}.
\end{abstract}


\begin{figure*}[h]
    \centering
    \includegraphics[width=0.8\linewidth]{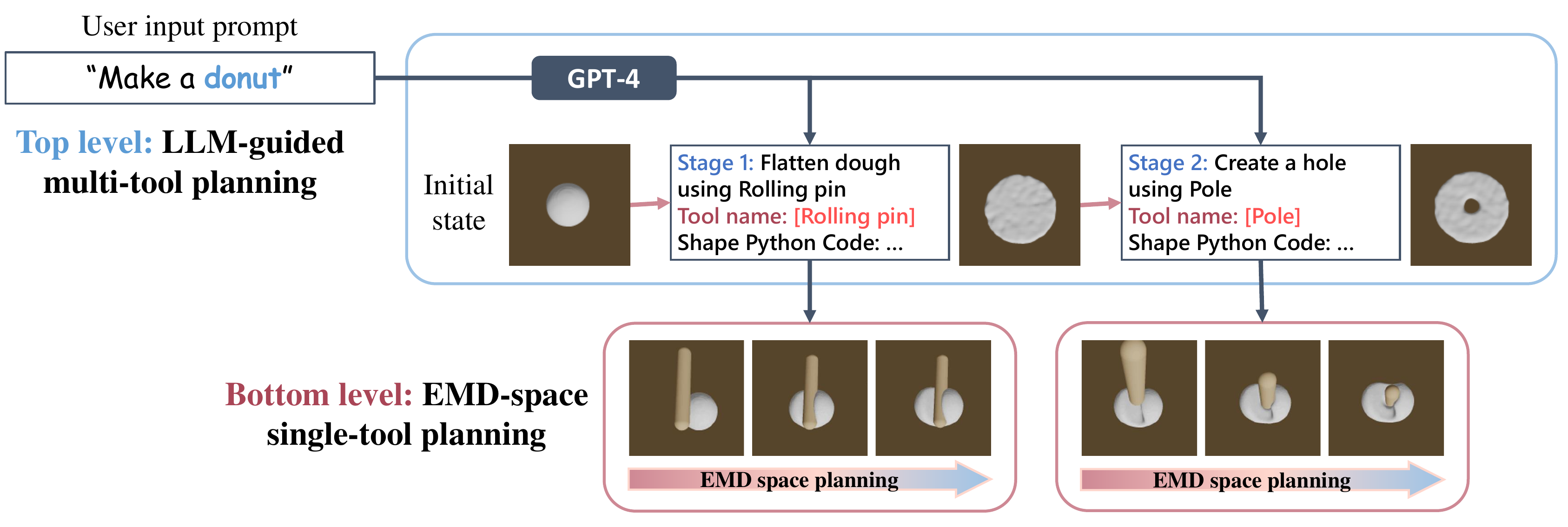}
    \caption{Schematic illustration of our method in handling unseen dough making tasks, where Language Models (LLMs) are utilized at a high level for task decomposition and subgoal generation, specifying tool names and generating corresponding Python code. The low-level operates on particle space controls, precisely determining the next achievable candidate iteratively without the need for prior demonstrations or task-specific training.
    }
    \vspace{-1.5em}
    \label{fig:overview}
\end{figure*}

\section{Introduction}
\IEEEPARstart{M}{anipulation} of deformable objects remains one of the most intricate challenges in robotics due to the inherent complexity and unpredictable behavior of such objects. Deformable objects can be categorized into two major categories: thin-shell surfaces, like clothes~\cite{nocentini2022learning,wu2019learning} and ropes~\cite{sundaresan2020learning}; and volumetric objects, like dough~\cite{lin2022planning,huang2021plasticinelab}. In this paper, we focus on the latter and study dough manipulation given a set of candidate tools like rolling pin, knife, etc.

Existing works on dough-like volumetric deformable objects majorly rely on a learned dynamic model of the underlying particles ~\cite{zhu2022challenges,arriola2020modeling,yin2021modeling}. 
However, these methods all require a substantial amount of collected or semi-auto-generated demonstrations for training the dynamic models, which poses two critical issues: firstly, the difficulty of obtaining a comprehensive set of demonstrations, particularly for long-horizon tasks; and more importantly, the limited capability of generalizing beyond the provided demonstrations. 

Given this context, there is an imperative need for a more versatile and universally applicable approach to deformable object manipulation, one that can navigate the intricacies of both short and long-horizon tasks, without being overly reliant on demonstrations. This paper introduces a novel demonstration-free hierarchical planning method that addresses these challenges. 

In this study, we delve into the manipulation of dough, a quintessential example of deformable object manipulation~\cite{lin2022planning,lin2022diffskill,huang2021plasticinelab} with tools. As illustrated in Figure \ref{fig:overview}, our approach takes a natural language user prompt as input and leverages a \highlight{LLM} to formulate a high-level plan detailing the selection of tools and the representation of intermediate subgoals at each phase. While LLMs might not produce precise low-level actions for each timestep, they exhibit proficiency in breaking down intricate tasks into manageable stages. Each of these stages exclusively involves a single tool and a piece of dough.

The concept of anchoring language to a sequential plan has been investigated in prior research~\cite{huang2022language,ahn2022can,liang2023code}. However, such methodologies have largely been confined to generating high-level linguistic instructions for robots for generic household tasks (e.g., ``open the fridge'' or ``bring me the apple''). They haven't been tailored for intricate tasks like deformable object manipulation. Indeed, there is a significant gap in literature when it comes to utilizing LLMs for manipulating deformable entities, especially when the challenge entails crafting complex shapes (like donuts or baguettes) based purely on linguistic outputs. In our approach, rather than defining the robot's actions or policy linguistically at intermediate stages, we direct LLMs to express their object-centric state visualizations via Python code. This distinctive approach sets our method apart from previous techniques.

In bridging adjacent subgoal imaginations generated from LLMs, we introduce a simple but novel EMD space planning algorithm complemented by model predictive control. The algorithm evaluates the gradient of the earth mover's distance between the prevailing point cloud and the target point cloud concerning each discrete point. Subtracting this gradient from the current point cloud yields the succeeding viable candidate. This process facilitates a direct point-to-point correspondence between the existing state and the upcoming candidate, enabling the deployment of differentiable physics based on a straightforward per-point mean squared error.

Through this hierarchical strategy, our system can adeptly tackle novel, intricate tasks devoid of any demonstrations or prior training, given a set of candidate tools. Experimental results demonstrate that our methodology markedly enhances the efficacy of both single-tool and multiple-tool dough manipulation with various tools and can potentially transfer to real-world robotic applications. \highlight{To summarize, we propose a novel demonstration-free hierarchical planning method that tackles complex long-horizon deformable manipulation tasks without training. Our contributions are two-fold:
\begin{itemize}
    \item We leverage LLMs to manipulate dough by decomposing tasks into distinct stages with achievable subgoals.
    \item We employ Differentiable Physics with Point-to-Point correspondence to execute dough manipulation iteratively, eliminating the need for demonstrations.
\end{itemize}
    }
\section{Related Works}

\paragraph{Differentiable physics for deformable object manipulation}
Differentiable physics is a pivotal technique in deformable object manipulation. It exploits the gradient from differentiable simulators to derive optimal actions. Existing literature~\cite{hu2019difftaichi,hu2019chainqueen,liang2019differentiable} reveals that differentiable physics offers an efficient means to tackle short-horizon simple tasks. Nevertheless, as highlighted by~\cite{antonova2023rethinking}, the reliance of differentiable physics on local gradients poses challenges. The loss landscape is often rugged with potentially spurious local optima, making it less reliable for certain tasks.

\paragraph{Long-horizon planning for deformable object manipulation}
There's an emerging interest in long-horizon strategies for deformable object manipulation. DiffSkill~\cite{lin2022diffskill} employs a latent space planner to assess various skill combinations and sequences to achieve the desired outcome. Subsequently, PASTA~\cite{lin2022planning} introduced the PlAnning with Spatial-Temporal Abstraction framework, melding spatial and temporal abstraction for effective long-horizon task planning. Robocraft~\cite{shi2022robocraft} advances a particle-based dynamics model using graph neural networks (GNNs) to grasp the system's structural nuances. This knowledge is then harnessed to optimize action sequences from a randomly initialized trajectory. Robocook~\cite{shi2023robocook} adopts point cloud scene representation and leverages GNNs for modeling tool-object interactions. It then synergizes tool classification with self-supervised policy learning for crafting object manipulation plans. Nonetheless, these methodologies have their constraints. They necessitate prior insight into potential tool candidates and a predetermined number of stages, which affects their adaptability. 

\paragraph{Language models for robot manipulations}
Leveraging large language models for robotics is currently a bustling research domain. Recent studies such as~\cite{huang2022language,liang2023code,ahn2022can} strive to dissect complex tasks into manageable sub-stages. These methods, although innovative, are primarily innocent to the underlying geometry, providing only high-level robot directives. To enrich these models with diverse modalities, SM~\cite{zeng2022socratic} developed a modular framework where new tasks are delineated as a linguistic interaction between pre-trained models and auxiliary modules, underpinned by Socratic reasoning. PaLM-E~\cite{driess2023palm} engineered a versatile embodied multimodal language model tailored for a spectrum of downstream reasoning challenges. VoxPoser~\cite{huang2023voxposer} harnesses LLMs to craft voxel value maps in a 3D observation space, guiding robot-environment interactions. Since LLMs often cannot directly produce the robot's raw actions, an alternative approach is to map intricate tasks to rewards. 
Some other studies~\cite{goyal2019using,lin2022inferring} focus on curating domain-specific reward models, which necessitate abundant annotated training data. In contrast, works like~\cite{kwon2023reward} generate reward metrics automatically from pretrained LLMs, though their application is predominantly limited to rigid or articulated objects. Deformable object manipulations remain a relatively under-explored territory for LLMs, largely due to the immense degrees of freedom inherent to such tasks and the paucity of available demonstration data.

\section{Method}
Given a set of candidate common tools to accomplish a complex dough manipulation task, our method adopts a hierarchical planning approach combining both language models and low-level particle space controls. At the top level, LLMs are employed to break down a complex task into sub-stages, and output both the code to generate subgoal states and the tool name for each. We observe that LLMs obtain rich knowledge about high-level task semantics though cannot directly output raw low-level actions.
At the bottom level, given the current tool and subgoal, our technique iteratively identifies the next reachable target based on the present state and subgoal. A key distinction between our approach and previous ones is that ours doesn't necessitate any demonstration or training for the target task. In Section~\ref{sec:multiplanning}, we detail the partitioning of complex tasks into sub-stages and the associated tools. Section~\ref{sec:singleplanning} elaborates on the iterative process of determining the next goal based on the current state and subgoal.

An overview of our method is illustrated in Figure~\ref{fig:overview}. Our method processes the sampled particles from the volumetric dough as input and produces the actions of the currently used tool as output. In the subsequent context, we will interchangeably use point clouds and particles.
\subsection{Multiple Tool Selection and Hierarchical Planning}
\label{sec:multiplanning}

To address the challenge of coordinating between different tools in long-horizon tasks, we turn to the capabilities of \highlight{LLMs} especially ChatGPT-4~\cite{openai2023gpt4}. Our observations indicate that while LLMs may not precisely produce low-level actions, they excel in deconstructing intricate long-horizon tasks into several stages, each centered around a single tool. What's more, for each of these stages, the LLM can both identify the appropriate tool and generate the corresponding Python code to produce intermediate subgoal point clouds. Presented in the form of particles, these subgoals readily align with the target requirements of our proposed single-tool EMD space planning algorithm (Section \ref{sec:singleplanning}). 

To guide this process, we devised a prompt template that imparts to the LLM foundational information about available tools and their potential interactions with the dough, \highlight{which is shared by all tasks. In this work, we consider four different tools: \textit{rolling pin}, \textit{gripper}, \textit{knife} and \textit{pole}. We require that only one tool be used per stage, which helps ensure that the LLM generates feasible targets matched to the chosen tool.} Additionally, we introduce a set of guidelines designed to refine and direct the LLM when completing more complex, long-horizon deformable tasks. One important guideline is to force the LLM to give volume-preserving input and output at each stage, so the target is more physically realistic. In addition, we leverage the chain reasoning technique ~\cite{wei2022chain} to help the LLM better deduce the shape parameters to satisfy this constraint. In Table~\ref{tab:volume_change}, we provide the average relative volume change between the LLM's generated final output and the input dough for all three evaluated tasks. \highlight{Importantly, we do not encode any task-specific or object-related information in our prompt, but instead rely on the LLM's grounding knowledge to select appropriate tools and decompose tasks. We can execute novel tasks by simply adding one line after our prompt. For instance, to make a donut, we only need to concatenate: ``the dough is initially a round ball at (0, 0, 0) with radius 1. Make a simple donut.''} Detailed prompt  can be found on our project website.

\begin{table}[h]
\caption{Volume change with and without volume-preserving (VP) and chain reasoning (CR).}
    \centering
    \vspace{-0.5em}
    \begin{tabular}{lccc}
    \toprule
    & Donut $\downarrow$& Baguette $\downarrow$& TwoPancakes$\downarrow$\\
            \midrule
        w/o VP and CR & 73.9\% & 42.5\% & 65.0\% \\
        Ours & \textbf{9.8\%} & \textbf{38.9\%} & \textbf{0.0\%} \\
        \bottomrule
    \end{tabular}
    
    \label{tab:volume_change}
\end{table}

In addition, we also ask the LLM to output the following items for each stage during planning:
\begin{itemize}
    \item A one-line explanation of what this step is doing.
    \item The name of the tool to be used. 
    \item The Python code to generate target point cloud. 
    \item The variable name for the generated point cloud.
    \item The location of each piece.
    \item The volume of each piece. 
\end{itemize}

Building on this, we only extract the generated Python code corresponding to each stage, \highlight{where the LLM utilizes the \textit{numpy} library to construct and save the subgoal point cloud into a variable}, from which we produce the intermediate subgoal point cloud. Other outputs, though not used, are part of chain reasoning and, therefore, contribute to the final quality of the generated subgoal. The subgoal point cloud is then fed directly into our single-tool planning module. Consequently, we are able to tackle intricate tasks incrementally, stage by stage, without any demonstrations.

\subsection{Single Tool Planning}
\label{sec:singleplanning}

As the LLM can decompose a complex task into several stages with generated sub-targets, on each stage,
given the current point cloud and the sub-target, we introduce a novel closed loop planning algorithm.

Our method initially identifies an optimal starting position for the tool. Subsequently, it forecasts the nearest attainable target and refines the actions employing differentiable physics with point-to-point correspondences, termed DiffPhysics-P2P. If this step doesn't yield progress, our model reverts the actions and re-strategizes using a new starting position. This process repeats for several steps until we reach the target goal or hits the maximum number of steps. The algorithm is summarized in Algorithm~\ref{alg:all}. 

\paragraph{DiffPhysics-P2P}
Given the current observation and the goal, we pinpoint the subsequent reachable point cloud by executing multiple small iterations (specifically 20 in our experiments) of gradient descent \highlight{with step size $\alpha$ } in the Earth Mover Distance space. Formally, each iteration is:
\begin{equation}
\label{eq:emd}
    \mathbf{p}_i' = \mathbf{p}_i - \alpha\cdot\frac{\partial \mathrm{emd}(\{\mathbf{p}_i\}, \{\bar{\mathbf{p}}_i\})}{\partial \mathbf{p}_i}.
\end{equation}

\begin{figure}[t]
    \centering
    \includegraphics[width=\linewidth]{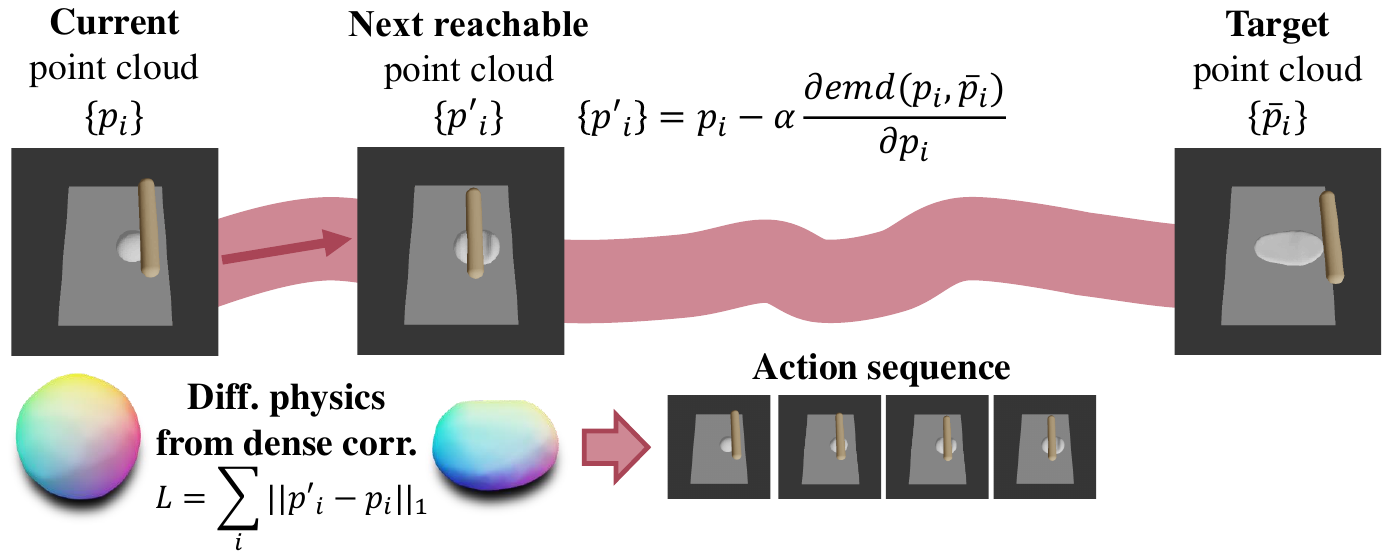}
    \caption{EMD-space planning with DiffPhysics-P2P. We find the next reachable target by running small steps within the EMD space. The induced point-to-point correspondence can provide better gradients when optimizing actions.
    }
    \label{fig:singleplanning}
\end{figure}

\highlight{EMD, also known as norm-1 Wasserstein Distance, measures the distribution difference between two point clouds. In our implementation, we compute it with Sinkhorn divergence approximation using GeomLoss~\cite{feydy2020geomloss} Python library.}
Given the current point cloud observation, denoted as ${\mathbf{p}_i}$, where $i$ is the point index, and the subgoal ${\bar{\mathbf{p}}_i}$, our objective is to discern the ensuing reachable candidate. This is achieved by incrementally transitioning the current point cloud towards the target within the EMD space. The candidate serves as our model's prediction of the underlying particle dynamics.

A notable advantage of the EMD gradient is its inherent capacity to furnish a one-to-one correspondence between $\mathbf{p}_i'$ and $\mathbf{p}_i$, as elucidated by Equation~\ref{eq:emd}. This characteristic permits the application of the subsequent straightforward mean-absolute-error loss:
\begin{equation}
    \mathcal{L} = \sum_i\|\mathbf{p}_i' - \mathbf{p}_i\|_1.
\end{equation}

This is different from several preceding methodologies, wherein the naive EMD loss (expressed as $ \mathrm{emd}({\mathbf{p}_i}, {\mathbf{p}_i'})$) is employed devoid of any point-to-point correspondence. Ablation studies underscore that our point-to-point correspondence substantially outperforms traditional differential physics by enhancing the gradient flow. An illustration of this algorithm is given in Figure~\ref{fig:singleplanning}.


\paragraph{Initial position selection}
At the beginning of each stage, we use the strategy similar to that in~\cite{li2022contact} to figure out a good initial position for the tool. Specifically, with the induced point-to-point correspondence (represented as $\mathbf{p}_i' - \mathbf{p}_i$), we find the best tool position $\mathbf{x}^*$ by:
\begin{equation}
 \mathbf{x}^* = \arg\max_\mathbf{x} (\sum_i \frac{\|\mathbf{p}_i' - \mathbf{p}_i\|_1}{d(\mathbf{p}_i,\mathbf{x}) + \delta}).
\end{equation}
The denominator computes the signed distance between point $\mathbf{p}_i$ to the closest point on the tool
placed at $\mathbf{x}$ (plus small $\delta$ to avoid division by zero). \highlight{Intuitively, this summation requires positioning the tool at the location where significant deformation is about to happen.
}


\paragraph{Tool reset upon failures}
\begin{figure}[t]
    \centering
    \includegraphics[width=\linewidth]{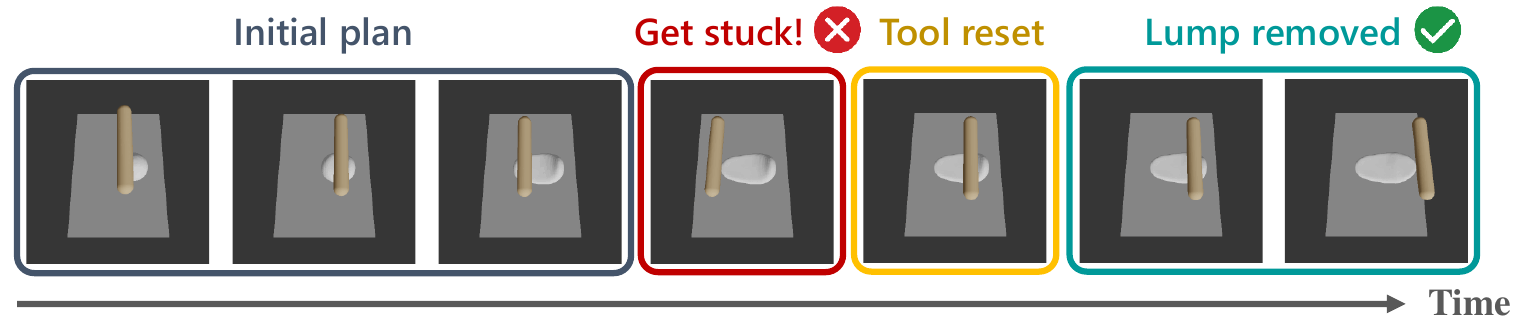}
    \caption{Illustration on how tool reset works. By resetting the tool position when no improvement can be made, we can jump out of the local minima.
    }
    \label{fig:tool_reset}
\end{figure}

As illustrated in Figure~\ref{fig:tool_reset}, consider a scenario where the task is to use a rolling pin to flatten the dough. Sometimes our planning algorithm will get stuck without further progress, leaving a lump. We find that resetting the tool's position when there is no advancement in the EMD loss helps escape from potential local minima.

The comprehensive single-tool closed loop planning process is detailed in Algorithm~\ref{alg:all}.

\begin{algorithm}[ht!] 
\caption{Closed Loop Planning with Model Predictive Control}
\label{alg:all}
\begin{algorithmic}[1] 
\Require Current system particles $\{\mathbf{p}_i\}$, target particles $\{\bar{\mathbf{p}}_i\}$
\Ensure Predicted actions at each timestep $\{\mathbf{a}_t\}$
\State $t:=0, need\_reset := 1, \mathrm{emd}_{last} := \infty$
\While{$t < max\_steps$}
    \State $\{\mathbf{p}_i'\} := \{\mathbf{p}_i\}$
    \For{$k$ \textbf{in} $1...K$}
        \State $\{\mathbf{p}_i'\} := \{\mathbf{p}_i'\} - \alpha\cdot\nabla_{\{\mathbf{p}_i'\}}\mathrm{emd}(\{\mathbf{p}_i'\}, \{\bar{\mathbf{p}}_i\})$ 
    \EndFor
    \State Set $\{\mathbf{p}_i'\}$ as the next reachable candidate
    \If{$need\_reset$}
        \State $\mathbf{x}^* = \argmax_\mathbf{x} \frac{\|\mathbf{p}_i' - \mathbf{p}_i\|_1}{\mathrm{sdf}_\mathbf{x}(\mathbf{p}_i) + \delta}$ \
        \State Set initial tool position to $\mathbf{x}^*$
        \State $need\_reset=0$
    \EndIf
    \State $\mathbf{a}_{t:t+L} := \mathbf{0}$ 
    \For{$j$ \textbf{in} $1...J$}
        \State $\mathbf{a}_{t:t+L} := \mathbf{a}_{t:t+L} - \nabla_{\mathbf{a}}\mathrm{Sim}(\sum_i \|\mathbf{p}_i' - \mathbf{p}_i\|_1)$ 
    \EndFor
    \State Execute $\mathbf{a}_{t:t+L}$ and get new particle observation $\{\tilde{\mathbf{p}}_i\}$ 
    \State $\mathrm{emd}_{curr} := \mathrm{emd}(\{\tilde{\mathbf{p}}_i\}, \{\bar{\mathbf{p}}_i\})$ 
    \If{$\mathrm{emd}_{curr} > \mathrm{emd}_{last}$} 
        \State $need\_reset = 1$ 
        \State \textbf{continue}
    \EndIf
    \If{$\mathrm{emd}_{curr} < \tau$} 
        \State \textbf{break}
    \EndIf
    \State $t = t+L, \mathrm{emd}_{last}=\mathrm{emd}_{curr}, \{\mathbf{p}_i\}=\{\tilde{\mathbf{p}}_i\}$ 
\EndWhile
\end{algorithmic}
\end{algorithm}

\highlight{
From L4 to L6, we determine the next reachable point cloud by moving particles along the gradient in the EMD space for \textit{K} iterations. At L9, we compute the optimal initial tool position. Between L13 and L16, we optimize a sequence of actions with a horizon \textit{L} by running differentiable physics and backpropagating gradients to the actions. Subsequently, we execute the actions and compute the EMD to measure the deviation from the target at L18. If no progress is made (L19), we reset the tool position (L20). At L23, if the system is sufficiently close to the target, we break; otherwise, we continue until the maximum number of timesteps (L2). The model predictive control (MPC) lies in that our model first predicts $\mathbf{p}_i'$ as the next reachable candidate (L7 of Algorithm 1) — this is the model predictive part, and then uses that prediction to optimize the actions (L15 of Algorithm 1) — this is the optimize and control part
}

\subsection{Implementation Details on Simulated Experiments}
\label{sec:details}
Our simulation environments are in line with established dough manipulation literature, notably Lin et al.~\cite{lin2022diffskill,lin2022planning}. We employ both DiffTaichi~\cite{hu2019difftaichi} and PlastineLab~\cite{huang2021plasticinelab} to handle differentiable physics. Following the common practice, we combine three distinct losses with coefficient $(1, 1, 0.02)$ when optimizing actions with differentiable physics:

\begin{enumerate}
    \item The point-to-point loss described in Section~\ref{sec:singletool}.
    \item An SDF loss that make the tool close to the dough.
    \item A velocity regularization loss.
\end{enumerate}

We use a parameter \(max\_steps=200\), \(K=20\) and \(L=40\). We make use of the Adam optimizer with a learning rate set at 0.02 during the differentiable physics computations. 
\section{Experiment}

In our experiments, we validate the efficacy of our method across four distinct dimensions. In Section~\ref{sec:multitool}, we deploy our hierarchical LLM-guided planning algorithm to more intricate and unseen long-horizon tasks. In Section~\ref{sec:singletool}, we conduct simpler tasks involving only one tool to authenticate the effectiveness of our single-tool planning algorithm. In Section~\ref{sec:ablation}, we perform distinct ablation studies for both single-tool and multiple-tool planning algorithms. 
In Section~\ref{sec:real}, we translate our simulated actions to a real-world robot to manipulate the actual dough, illustrating the practical applicability of our method from sim to real. 

\paragraph{Baselines}
We employ the following baselines for comparisons: Firstly, we consider a simplistic gradient-based action optimization method utilizing differentiable physics, denoted as Diff. Physics. Secondly, we examine a sophisticated long-horizon planning algorithm, PASTA~\cite{lin2022planning}, which integrates both spatial and temporal abstraction. Thirdly, we explore a behavior cloning method that trains a goal-conditioned policy, abbreviated as BC. Lastly, we assess a model-free RL method, SAC-N~\cite{an2021uncertainty}.
For complex tasks, generating a large dataset of demonstrations is intractable. Thus we manually annotate a single demonstration sequence for each task for training BC and SAC-N. For PASTA, we employ its pre-trained model on single-tool demonstrations and assess its generalization capabilities on these unseen, intricate tasks.
All the compared methods receive the point cloud of particles as input.

\paragraph{Metrics}
We adopt the metrics in~\cite{lin2022diffskill,lin2022planning} and report the normalized decrease (score) in EMD, which is computed  as \(\frac{\mathrm{emd}(\{\mathbf{p}_0\},\{\mathbf{p}^*\}) - \mathrm{emd}(\{\mathbf{p}_T\},\{\mathbf{p}^*\})}{\mathrm{emd}(\{\mathbf{p}_0\},\{\mathbf{p}^*\})}\), where \(\{\mathbf{p}_0\}\) is the initial point cloud, \(\{\mathbf{p}_T\}\) is the final point cloud after execution, and \(\{\mathbf{p}^*\}\) is the ground-truth target 3D point cloud.
Additionally, we also calculate the success rates based on pre-set thresholds (Appendix~\ref{sec:detail_desc}).

For each multiple-tool task, we utilize the LLM's Python code output from the last stage to generate the point cloud, serving as the ground truth. We observe that these point clouds, despite being generated by the LLM, are of high quality and adeptly describe the target shape. For single-tool tasks, we follow previous literature~\cite{lin2022planning,lin2022diffskill} to sample 5 random targets at different shapes and locations. Samples of the generated target point clouds are shown in Appendix~\ref{sec:target_point_clouds}.
For each multiple-tool task, we generate 4 distinct responses from the LLM, with each response being assessed 5 times, culminating in a total of 20 trials per task; for each single-tool task, we follow previous literature to evaluate one trial per target, resulting in 5 trials per task.

\subsection{Multiple-Tool Selection and Planning}
\label{sec:multitool}

\paragraph{Environment setup} We examine three intricate long-horizon tasks that necessitate the use of multiple tools: Donut, Baguette, and TwoPancakes. As implied by their names, these tasks require the agent to create a donut, a baguette, and two pancakes, respectively.
In the TwoPancakes task, the dough is initially presented as a rectangle. For the Donut and Baguette tasks, the dough is initially provided in the form of a unit ball.
A more comprehensive description of each task can be found in Appendix~\ref{sec:detail_desc}.

\vspace{-0.5em}
\begin{figure}[ht]
    \centering
    \includegraphics[width=\linewidth]{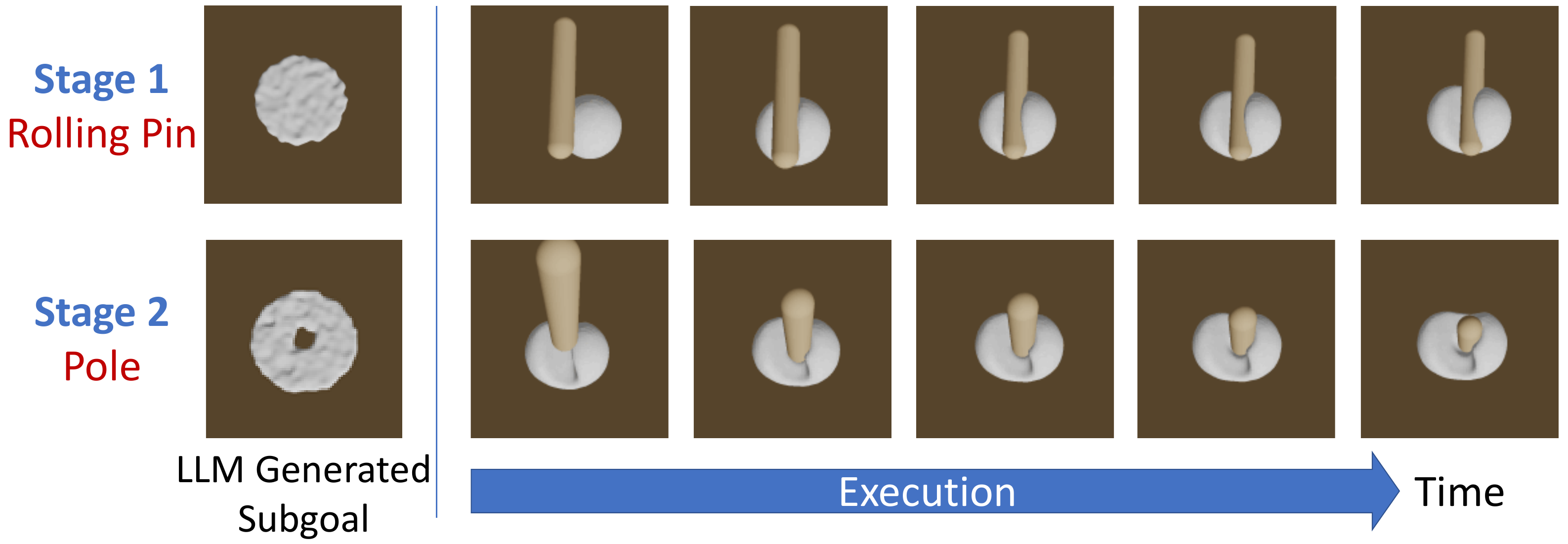}
    \vspace{-1.5em}
    \caption{``Make a Donut.'' An exemplary zero-shot execution on complex long-horizon tasks.}
    \label{fig:multitool}
\end{figure}

\begin{figure}[ht]
    \centering
    \vspace{-1.5em}
    \includegraphics[width=\linewidth]{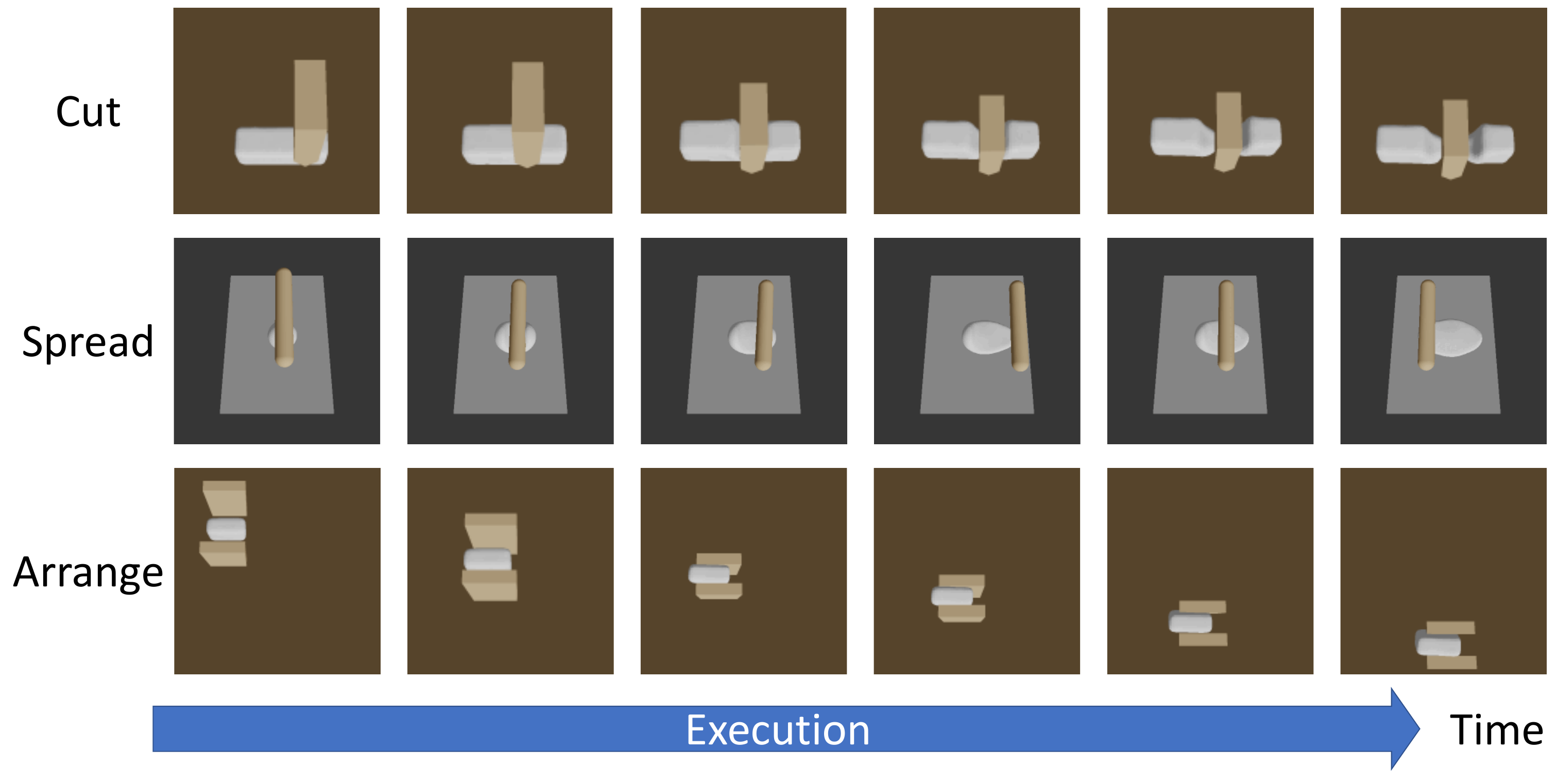}
    \vspace{-1.5em}
    \caption{Cut, spread, arrange. Single-tool execution results.}
    \label{fig:singletool}
\end{figure}
\vspace{-0.5em}

\paragraph{Results}
The quantitative results are presented in Table~\ref{tab:results}'s left. It is evident from the data that our method substantially surpasses preceding approaches, exhibiting superior performance across all three tasks by a considerable margin. It is crucial to underscore that our model has never been exposed to these tasks before, and it employs the high-level stage plans generated by the LLM and the single-tool EMD space planning method to dynamically generate actions. A detailed, stage-by-stage visual representation of the process is provided in Figure~\ref{fig:multitool}, illustrating the nuanced steps and strategies employed by our method in navigating and accomplishing the tasks. 
More qualitative results are given in Appendix~\ref{sec:subgoal}.

The key distinction of our approach lies in its zero-shot learning ability, which enables it to adapt to novel tasks without task-specific fine-tuning or training. This is a significant leap over the sampled-based methods, which may not provide a feasible path or require extensive data for complex shapes. The \highlight{LLM} plays a crucial role in our framework by charting a high-level planning path, which serves as a guide for the subsequent execution by the low-level \highlight{EMD} space planning algorithm. This hierarchical structure is pivotal; the LLM alone cannot translate its generated plans into the raw actions required for robotic manipulation. Conversely, without the strategic direction provided by the LLM, the EMD space planning lacks a coherent objective, struggling to discern what end states are physically plausible for the robot to achieve. 
PASTA, while effective within its demonstrated scope, requires a dataset to train on in order to sample feasible intermediate states and is thus inherently limited in its ability to generalize to new shapes, e.g., donut. All their modules like VAE, cost predictor, etc., are tailored to their collected training data.  This data-driven dependency hinders its application to the more complex tasks in this paper.

\begin{table}[h]
\caption{Quantitative comparisons on both single-tool and multiple-tool tasks.}
\centering
\resizebox{\linewidth}{!}{
\begin{tabular}{lcccccc}
    \toprule
    \multirow{2}{*}{\large{Method}} & \multicolumn{3}{c}{\large{\textbf{Multiple-tool tasks}}} & \multicolumn{3}{c}{\large{\textbf{Single-tool tasks}}}\\
    \cmidrule(lr){2-4}\cmidrule(lr){5-7}
    ~ & \large{Donut} & \large{Baguette} & \large{TwoPancakes} & \large{Spread} &
    \large{Cut} & \large{Arrange} \\
    \midrule
    \normalsize{Diff. Physics}~\cite{hu2019chainqueen} & \normalsize{0.141/0\%} & \normalsize{0.175/20\%} & \normalsize{0.583/0\%} & \normalsize{0.184/20\%} & \normalsize{0.401/60\%} & \normalsize{0.296/20\%} \\
    \normalsize{PASTA}~\cite{lin2022planning} & \normalsize{0.020/0\%} & \normalsize{-0.116/0\%} & \normalsize{-0.856/0\%} & \normalsize{0.155/20\%} & \normalsize{0.060/40\%} & \normalsize{0.052/0\%} \\
    \normalsize{BC} & \normalsize{0.001/0\%} & \normalsize{-0.606/0\%} & \normalsize{0.220/0\%} & \normalsize{0.441/60\%} & \normalsize{-0.488/20\%} & \normalsize{-0.512/0\%} \\
    \normalsize{SAC-N}~\cite{an2021uncertainty} & \normalsize{0.003/0\%} & \normalsize{-0.306/0\%} & \normalsize{0.127/0\%} & \normalsize{0.000/0\%} & \normalsize{-2.827/0\%} & \normalsize{0.267/0\%} \\
    \midrule
    \normalsize{Ours} & \normalsize{\textbf{0.346}/\textbf{75\%}} & \normalsize{\textbf{0.501}/\textbf{75\%}} & \normalsize{\textbf{0.858}/\textbf{65\%}} & \normalsize{\textbf{0.680}/\textbf{100\%}} & \normalsize{\textbf{0.685}/\textbf{100\%}} & \normalsize{\textbf{0.981}/\textbf{100\%}} \\
    \bottomrule
\end{tabular}
}

\vspace{-1.5em}
\label{tab:results}
\end{table}

\subsection{Single-Tool Planning}
\label{sec:singletool}

\paragraph{Environment setup} In the simulation environment provided by PASTA~\cite{lin2022planning}, we examine three elementary tasks related to dough manipulation: Spread, Cut, and Arrange. Each of these tasks necessitates the use of only one tool at a time and can be finished within 200 time steps. More task descriptions can be found in Appendix~\ref{sec:detail_desc}.

\paragraph{Results} 
Quantitative outcomes are presented in Table~\ref{tab:results}'s right. It is evident that our approach consistently surpasses previous baselines by a substantial margin. The baselines fail to accomplish these tasks, whereas our method attains a 100\% success rate in all of them. It is also noteworthy that, in contrast to prior learning-based approaches, our method does not necessitate any demonstration data. Remarkably, our method does not even entail any training, rendering it immediately applicable to new tasks. Qualitative results are illustrated in Figure~\ref{fig:singletool}.

\vspace{-0.5em}
\subsection{Ablation Studies}
\label{sec:ablation}

\begin{figure}[ht]
    \centering
    \includegraphics[width=0.7\linewidth]{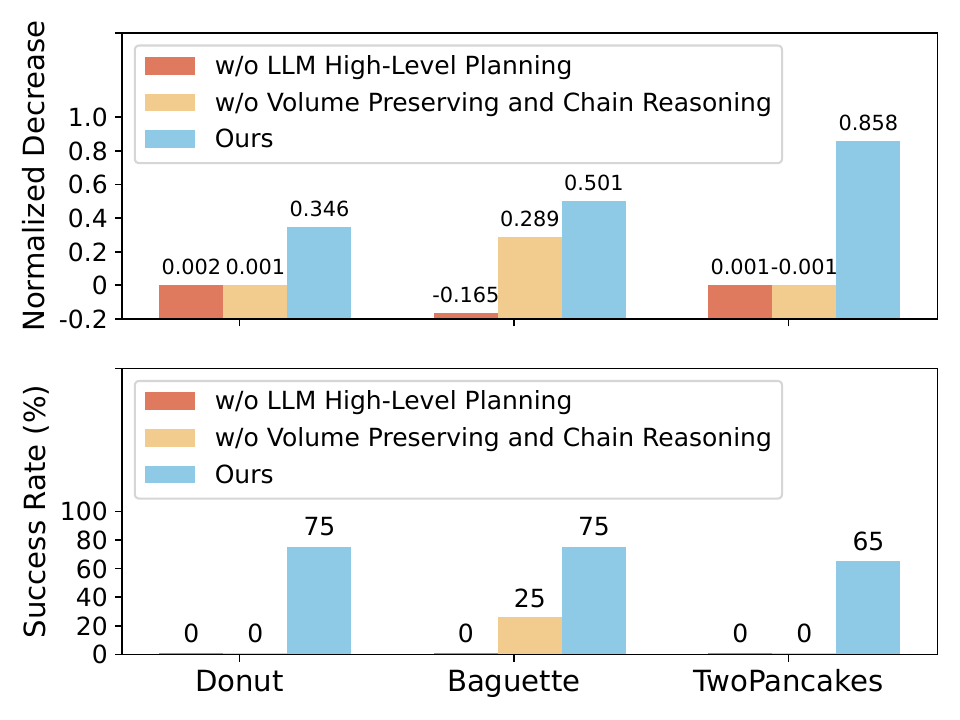}
    \vspace{-1em}
    \caption{Ablation results for multiple-tool experiments.}
    \label{fig:multitool_ablation}
\end{figure}

\paragraph{Multiple-tool ablations}
Figure~\ref{fig:multitool_ablation} presents our ablation studies on planning without the incorporation of high-level plans generated by the LLM. Additionally, we conduct ablation on the volume-preserving guidance with chain reasoning, which is proved to be crucial for maintaining both a high success rate and consistent target volume generation.

\begin{figure}[ht]
    \centering
    \includegraphics[width=0.7\linewidth]{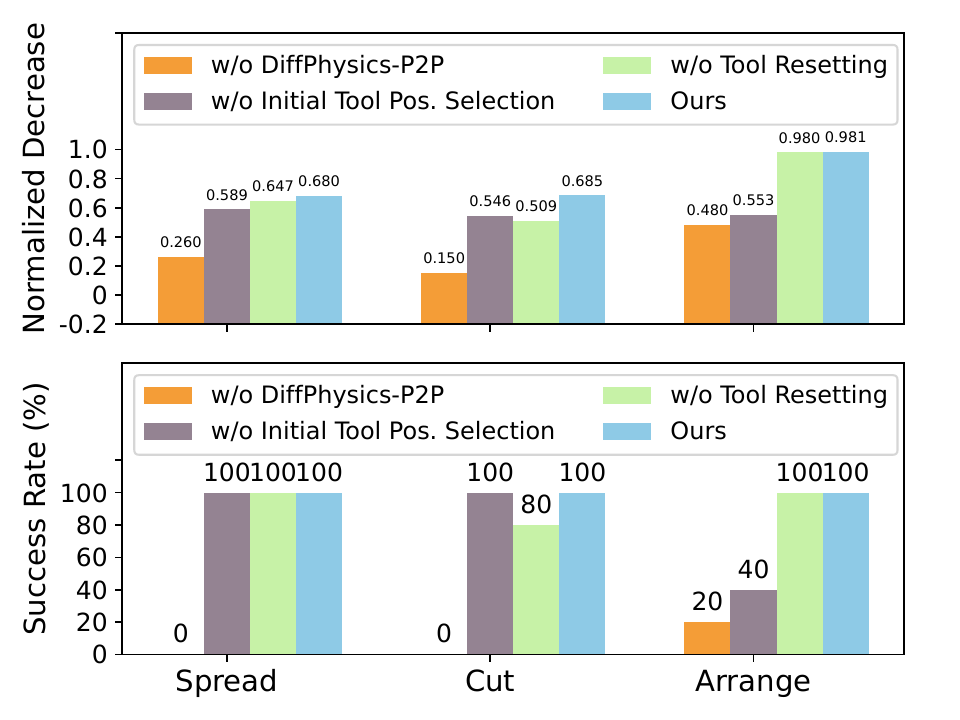}
    \vspace{-1em}
    \caption{Ablation results for single-tool experiments.}
    \vspace{-1.5em}
    \label{fig:singletool_ablation}
\end{figure}

\paragraph{Single-tool ablations}
Figure~\ref{fig:singletool_ablation} presents our ablation studies on the removal of the DiffPhysics-P2P, the initial position selection component, or the tool resetting module within the single-tool planning algorithm. \highlight{Without DiffPhysics-P2P means the system directly calculates the EMD loss between $\mathbf{p}_i$ (the initial state for this stage) and $\bar{\mathbf{p}_i}$ (the LLM-generated goal for this stage), with no point-to-point correspondence. It then optimizes all actions in this stage using $ \mathbf{a} = \mathbf{a} - \nabla_\mathbf{a}\mathrm{sim}(\mathrm{emd}(\bar{\mathbf{p}}_i,\mathbf{p}_i))$. Our approach differs by gradually moving through the EMD space using P2P loss $\sum_i\|\mathbf{p}_i'-\mathbf{p}_i\|_2$.}


\highlight{
\subsection{Time Analysis}
Finding the next reachable candidate (L4-L6 of Algorithm \ref{alg:all}) takes only 54ms using modern GPUs. Each iteration of the action gradient step with horizon $L=40$ (L15 of Algorithm \ref{alg:all}) takes 0.67s, and we perform $J=20$ iterations (L14-L16 of Algorithm \ref{alg:all}). Therefore, the average planning time per action is 0.67 * 20 / 40 = 0.34s, or approximately 3Hz.
}
\subsection{Real-Robot Experiments}
\label{sec:real}
\begin{figure}[ht]
    \centering
    \includegraphics[width=\linewidth]{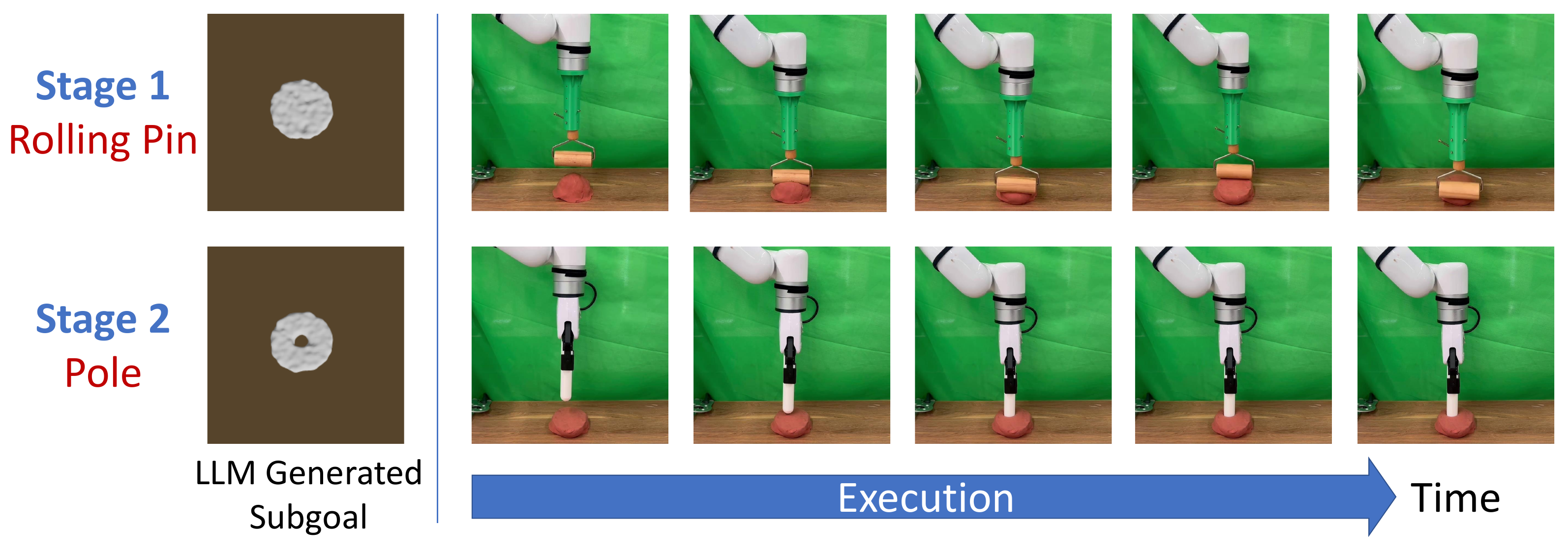}
    \caption{
    ``Make a donut.'' The robot first flattens the dough and uses the pole to create a hole.}
    \label{fig:realrobot}
\end{figure}

\paragraph{Environment setup}
We set up real-world experiments using the UFACTORY xArm 6 and some clay. Given the experimental setting and observations, the proposed planning algorithm is used to generate a trajectory in the simulator, and subsequently, the controller is employed to execute this trajectory in the real world. Unlike in simulation, we do not have the ground-truth volumetric particle state in real-world experiments. In order to estimate the particle from observations \highlight{and ensure the initial states in our simulation to be aligned with the real-world observation for each stage}, we set up four cameras from the four orthogonal directions and utilize truncated signed distance function (TSDF) fusion~\cite{zeng20163dmatch} to reconstruct volumetric particles from multi-view depth cameras. \highlight{During each stage, we measure the current particles in real world as the initial state and use LLM-generated subgoals as the target state in the simulator. Using our DiffPhysics-P2P method in the simulator, we gradually move from the initial to target state. We then transfer these simulated actions directly back to the real world.}
In Figure~\ref{fig:realrobot}, qualitative trajectories are provided for making a donut.

\begin{figure}[ht]
    \centering
    \includegraphics[width=0.95\linewidth]{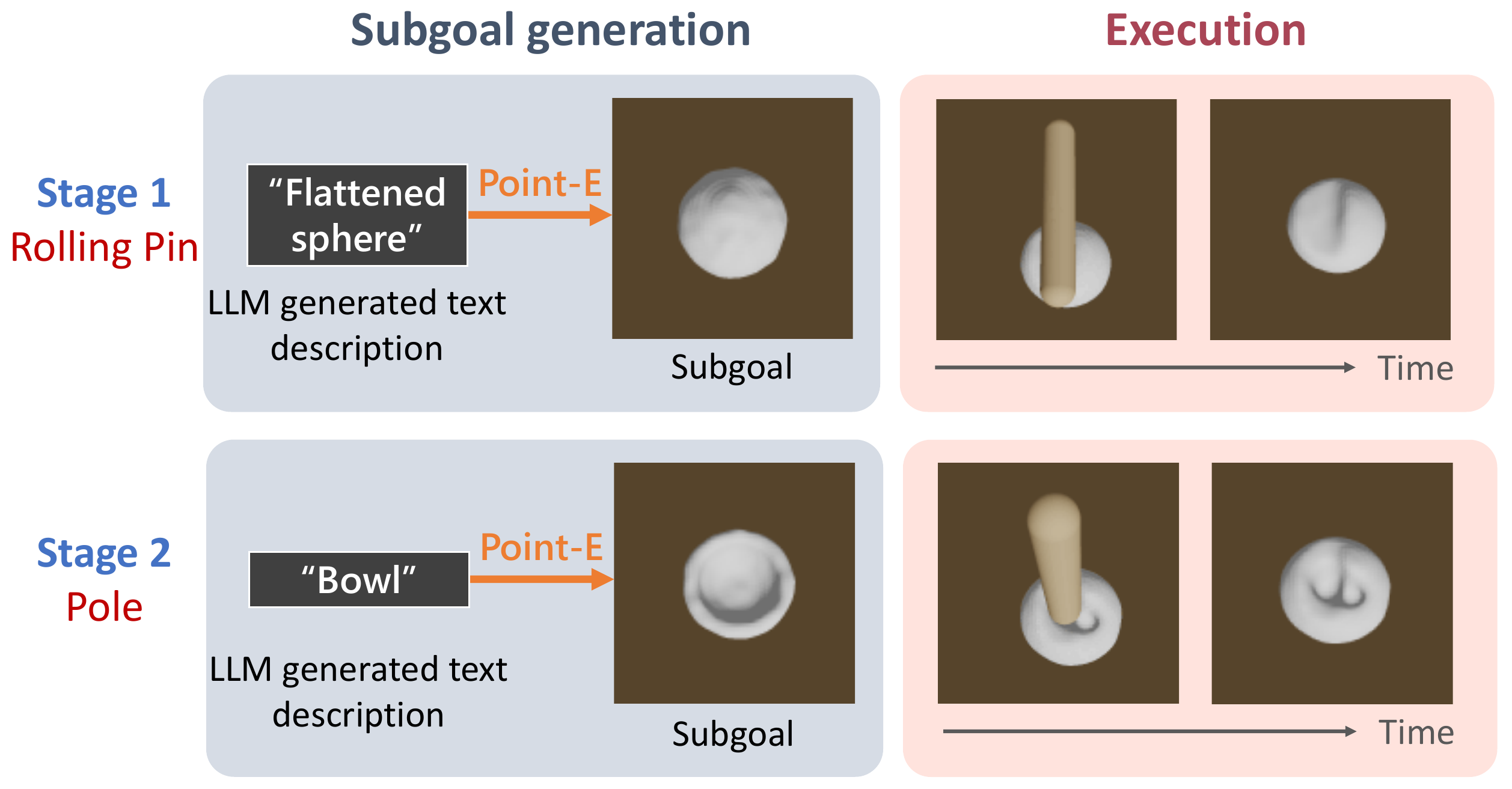}
    \caption{``Make a bowl.''
    An example that leverages off-the-shelf text-to-3D algorithms to generate complex shapes.
    }
    
    \label{fig:text3d}
\end{figure}
\section{Conclusions and Limitations}

We introduced a new hierarchical planning method for robotic deformable object manipulation, enabling complex tasks without prior demonstrations. 
Using large language models, it generates high-level plans and intermediate goals, which are executed through a unique closed-loop predictive control using Differentiable Physics. Our approach showed better performance and adaptability in dough manipulation benchmarks, marking a significant step forward in deformable object manipulation.

\paragraph{Limitations}
While our method is adept at handling complex tasks involving long-horizon planning, it is limited to generating simple shapes that can be described with the Python codes produced by the LLM. 
However, intermediate text descriptions can be converted by Text-to-3D generative models, such as Point-E~\cite{nichol2022point}. Figure~\ref{fig:text3d} illustrates an example of creating a bowl from the dough. The shape of a bowl is difficult to describe using pure Python code, so the LLM outputs text descriptions for each subgoal, like \textit{Flattened Sphere} and \textit{Bowl}. These texts are then interpreted by Point-E to generate point clouds. We then proceed with our planning, using these point clouds.


\appendix
\vspace{-0.5em}
\section{Appendix}
\subsection{Detailed Descriptions for each task}
\label{sec:detail_desc}
\paragraph{Single-Tool Tasks}

\begin{itemize}
    \item \textbf{Spread:} Spread is a task where the dough is initially a ball. The target is to use a rolling pin to flatten it into a flat sphere. The threshold for this task is 0.4.
    \item \textbf{Cut:} Cut is a task where the dough is initially an elongated box. The target is to cut the dough into two pieces in the middle with a knife. The threshold is 0.4.
    \item \textbf{Arrange:} Arrange is a task where the dough is initially a box. The target is to move the dough using a gripper to another place. The threshold is 0.7.
\end{itemize}
\paragraph{Multiple-Tool Tasks}
\begin{itemize}
    \item \textbf{Donut:} Donut is a task where the dough is initially a ball. The target is to create a donut-shaped dough. 
    The threshold for this task is 0.3.
    \item \textbf{Baguette:} Baguette is a task where the dough is initially a ball. The target is to create a baguette-shaped dough. 
    The threshold for this task is 0.5.
    \item \textbf{TwoPancakes:} TwoPancakes is a task where the dough is initially an elongated box. The target is to make two pancake-shaped doughs. 
    The threshold is 0.85.
\end{itemize}

\vspace{-0.5em}
\subsection{Hyperparameters for Simulation Dough}
We use PlasticineLab~\cite{huang2021plasticinelab} to simulate the dough. 
The physical parameters of the dough are set identical to those in PASTA~\cite{lin2022planning} to ensure a fair comparison.

\subsection{Target Point Clouds Visualization}
\label{sec:target_point_clouds}

\begin{figure}[t]
    \centering
    \includegraphics[width=\linewidth]{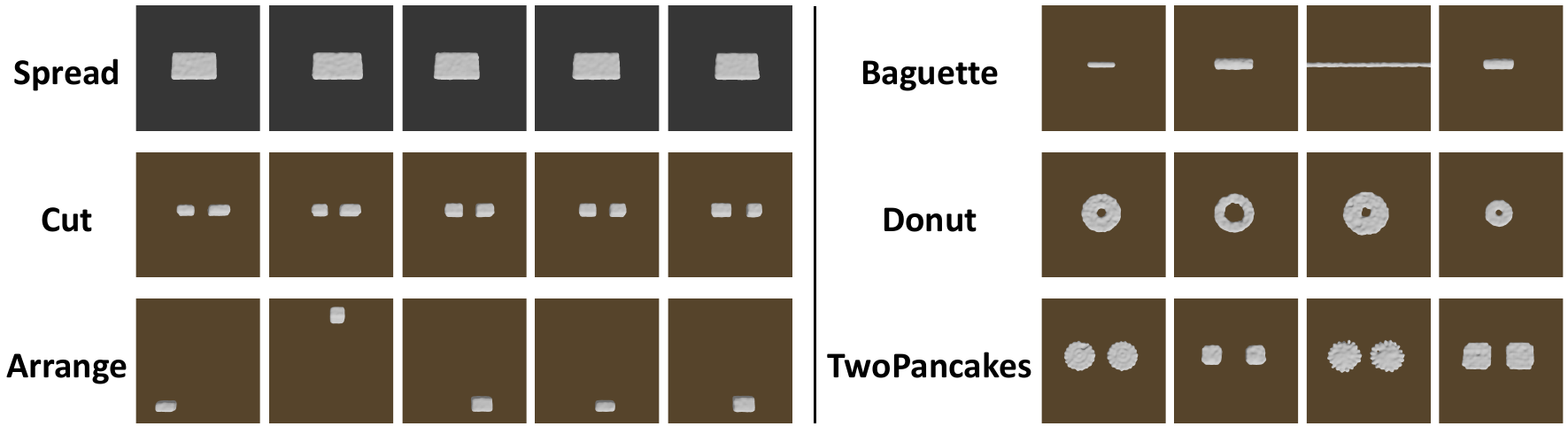}
    \caption{LLM-generated target point clouds.
    Left: Single-tool tasks.
    Right: Multi-tool tasks.
    }
    \label{fig:target_pointcloud}
\end{figure}

\paragraph{Single-tool tasks} For single-tool tasks, we follow PASTA~\cite{lin2022planning} to generate 5 random target dough point clouds at different places or shapes, as shown in Figure \ref{fig:target_pointcloud} left.
\paragraph{Multiple-tool tasks} For multiple-tool tasks, we utilize the LLM’s Python code output from the last stage to generate the point cloud, serving as the ground truth. We observe that these point clouds, despite being generated by the LLM, are of high quality and adeptly describe the target shape. Visualizations are illustrated in Figure \ref{fig:target_pointcloud} right.

\subsection{Intermediate Subgoal Visualization}
\label{sec:subgoal}
\begin{figure}
    \centering
    \includegraphics[width=\linewidth]{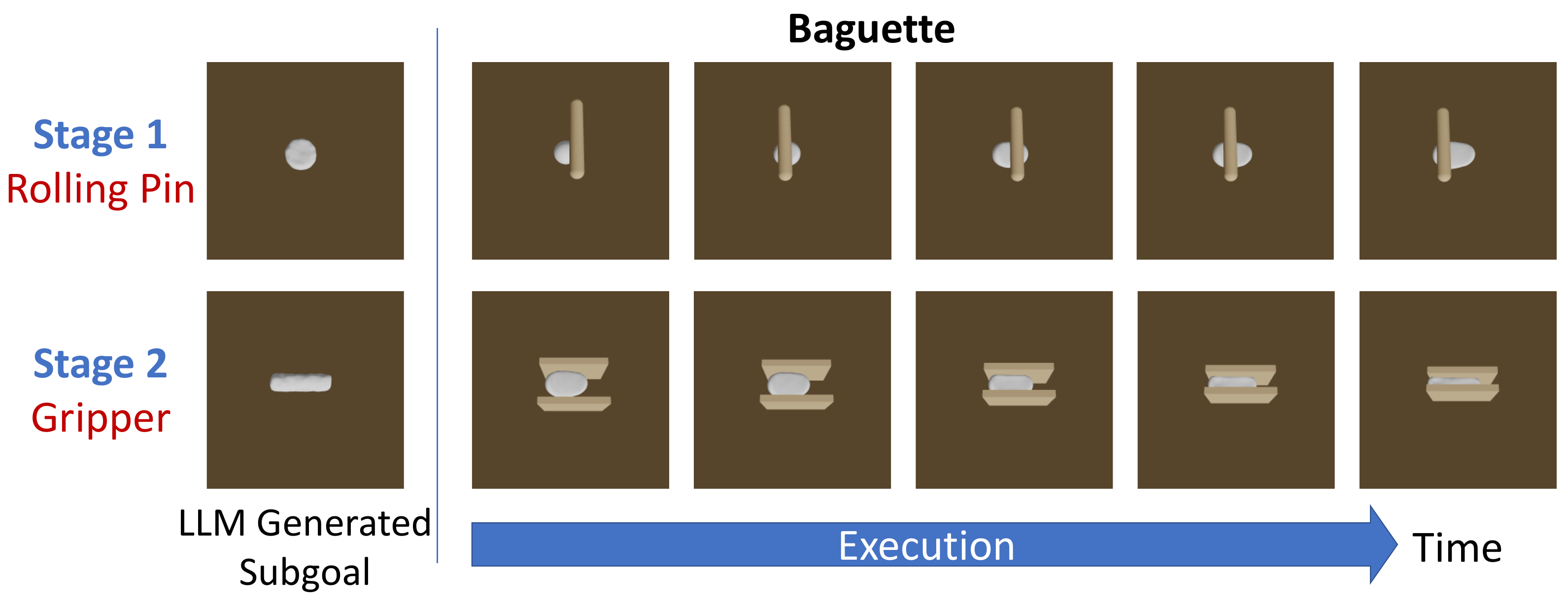}\\
    \includegraphics[width=\linewidth]{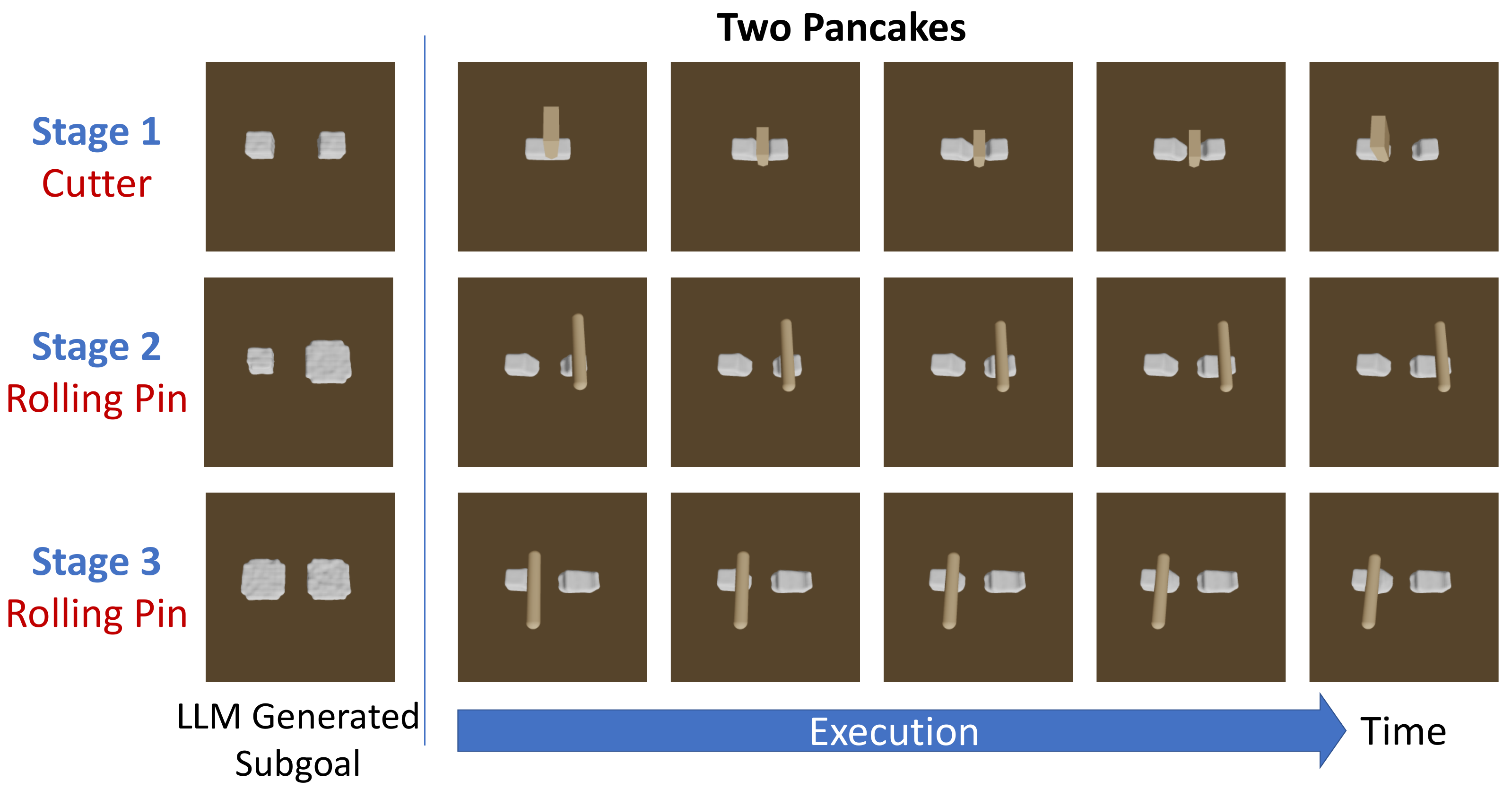}
    \caption{Execution trajectory of our method with intermediate subgoals set by the LLM.}
    \label{fig:ours_intermediate}
\end{figure}

Figure~\ref{fig:ours_intermediate} depicts more execution trajectories of our method, including intermediate goals generated by the LLM. While LLM may not produce subgoals that are perfectly reachable for execution, the closed-loop execution with EMD space point-to-point planning (DiffPhysics-P2P) algorithm within our approach exhibits resilience to such imperfections. 


\subsection{Next Reachable Point Cloud Visualization}
In this section, we visualize the topological change of next reachable point clouds throughout the process of making a donut.  The visualizations demonstrate key stages, such as the intentional alteration of the dough's topology to create a hole in the middle of a donut. 
Although our tools may not execute the point cloud to perfection, the LLM-generated subgoals guide the gradient flow towards the correct target. 

\begin{figure}[ht]
    \centering
    \includegraphics[width=\linewidth]{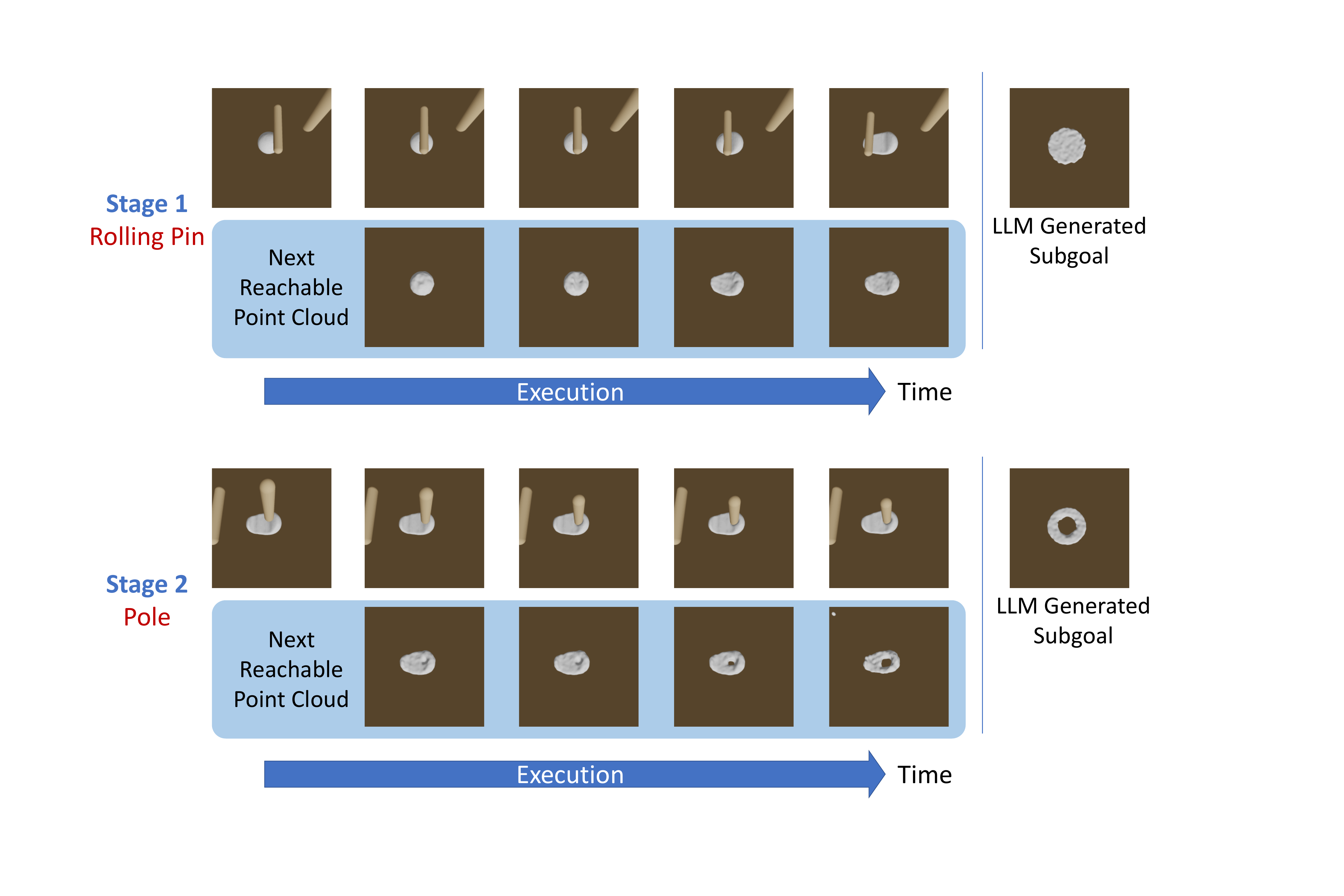}
    \caption{Next reachable point cloud visualization. 
    Notice that the topology is gradually changed during the second stage for making the donut. 
    LLM planning ensures that subsequent steps are guided towards physically realizable subgoals.
    }
    \label{fig:next_reachable}
\end{figure}

\bibliography{IEEEabrv,papers}
\bibliographystyle{IEEEtran}


\end{document}